%% file: root.tex
\documentclass[a4paper,conference]{IEEEtran}
\IEEEoverridecommandlockouts
\usepackage{url}

\usepackage{cite}
\ifCLASSINFOpdf
   \usepackage[pdftex]{graphicx}
\else
\fi
\usepackage{color}
\definecolor{green}{rgb}{0, 0.5, 0}
\definecolor{orange}{rgb}{1.0, 0.6, 0.2}
\definecolor{red}{rgb}{1.0, 0.0, 0.0}
\definecolor{teal}{rgb}{0.0, 0.4, 0.4}
\definecolor{purple}{rgb}{0.65,0,0.65}
\definecolor{saffron}{rgb}{0.95,0.75,0.2}
\definecolor{turquoise}{rgb}{0.0,0.5,0.5}
\definecolor{black}{rgb}{0.0, 0.0, 0.0}
\definecolor{gray}{rgb}{0.5, 0.5, 0.5}

\newcommand{\rui}[1]{{\color{purple}}}

\usepackage{amsmath}
\usepackage{algorithmic}
\usepackage{algorithm}
\usepackage{caption}
\usepackage[font=footnotesize]{subcaption}
\usepackage{tabularx}
\input{latex/common_cmds}

\begin{document}
% \title{GCS-CAM: Improving CNNs Visual Explanation by Grouped Channel Switching}
\title{FD-CAM: Improving Faithfulness and Discriminability  of Visual Explanation for CNNs}

% \author{\IEEEauthorblockN{Michael Shell}
% \IEEEauthorblockA{School of Electrical and\\Computer Engineering\\
% Georgia Institute of Technology\\
% Atlanta, Georgia 30332--0250\\
% Email: http://www.michaelshell.org/contact.html}
% \and
% \IEEEauthorblockN{Homer Simpson}
% \IEEEauthorblockA{Twentieth Century Fox\\
% Email: homer@thesimpsons.com}
% \and
% \IEEEauthorblockN{James Kirk\\ and Montgomery Scott}
% \IEEEauthorblockA{Starfleet Academy\\
% San Francisco, California 96678--2391\\
% Telephone: (800) 555--1212\\

\author{\IEEEauthorblockN{Hui Li, Zihao Li, Rui Ma$^*$, Tieru Wu$^*$ \thanks{$^*$ Corresponding authors. \newline This work is supported by the National Key Research and Development Program of China (Grant 2020YFA0714103) and the National Nature Science Foundation of China (Grant 61872162).}}
\IEEEauthorblockA{School of Artifical Intelligence \\
Jilin University, Changchun, China\\
lih20@mails.jlu.edu.cn, zihaol20@mails.jlu.edu.cn, ruim@jlu.edu.cn, wutr@jlu.edu.cn
}
}

%Telephone: (800) 555--1212, Fax: (888) 555--1212}

% \IEEEspecialpapernotice{(Invited Paper)}

\maketitle
% \footnote{This work is supported by the National Key Research and Development Program of China (Grant 2020YFA0714103) and the National Nature Science Foundation of China (Grant 61872162).}
%%%%%%%%% ABSTRACT
\input{latex/0_abstract}

% %%%%%%%%% BODY TEXT
\input{latex/1_introduction}

%%%%%%%%% RELATED WORK
\input{latex/2_related_work}
\input{latex/3_method}

% %%%%%%%%% EXPERIMENTS
\input{latex/4_experiments}
\input{latex/5_conclusion}

% \fi
\IEEEpeerreviewmaketitle

\newpage



\bibliographystyle{IEEEtran}
\bibliography{IEEEabrv}


% \end{thebibliography}

\end{document}

%% file: latex/common_cmds.tex
\usepackage{color}
\definecolor{green}{rgb}{0, 0.5, 0}
\definecolor{orange}{rgb}{0.8, 0.6, 0.2}
\definecolor{red}{rgb}{1.0, 0.0, 0.0}
\definecolor{teal}{rgb}{0.0, 0.4, 0.4}
\definecolor{purple}{rgb}{0.65,0,0.65}
\definecolor{saffron}{rgb}{0.95,0.75,0.2}
\definecolor{turquoise}{rgb}{0.0,0.5,0.5}
\definecolor{black}{rgb}{0.0, 0.0, 0.0}
\definecolor{gray}{rgb}{0.5, 0.5, 0.5}

%% file: latex/0_abstract.tex
\begin{abstract}

Class activation map (CAM) has been widely studied for visual explanation of the internal working mechanism of convolutional neural networks. 
The key of existing CAM-based methods is to compute effective weights to combine activation maps in the target convolution layer.
Existing gradient and score based weighting schemes have shown superiority in ensuring either the discriminability or faithfulness of the CAM, but they normally cannot excel in both properties.
In this paper, we propose a novel CAM weighting scheme, named FD-CAM, to improve both the faithfulness and discriminability of the CAM-based CNN visual explanation. 
First, we improve the faithfulness and discriminability of the score-based weights by performing a grouped channel switching operation. 
Specifically, for each channel, we compute its similarity group and switch the group of channels on or off simultaneously to compute changes in the class prediction score as the weights.
Then, we combine the improved score-based weights with the conventional gradient-based weights so that the discriminability of the final CAM can be further improved. 
We perform extensive comparisons with the state-of-the-art CAM algorithms.
The quantitative and qualitative results show our FD-CAM can produce more faithful and more discriminative visual explanations of the CNNs.
We also conduct experiments to verify the effectiveness of the proposed grouped channel switching and weight combination scheme on improving the results.
Our code is available at \url{https://github.com/crishhh1998/FD-CAM}.
\end{abstract}

%% file: latex/1_introduction.tex
\section{Introduction}
\label{sec:intro}

% In recent years, machine learning methods have received widespread attention for their remarkable results and its adoption has been expanding. Deep Neural Networks (DNNs), as one of the most widely used machine learning methods, have achieved great breakthroughs in various fields, especially in the domains of computer vision, which include  image classification\cite{krizhevsky2012imagenet,vgg},object detection\cite{he2017mask,dai2016r}, semantic segmentation\cite{badrinarayanan2017segnet,dai2016r} and so on. While DNNs still remain black boxs, their internal ambiguity is unquestionable. Therefore, the use of DNNs in many high-stakes domains is controversial such as healthcare and financial services. These extremely highlight the need of interpretability of DNNs which can help model deployers and users have more confidence in the results, thus decide whether to choose the model or whether to agree with the results of the model.

In recent years, deep learning has achieved great breakthroughs in various fields. 
Especially in the domain of computer vision, Convolutional Neural Networks (CNNs) have produced remarkable results for image classification \cite{krizhevsky2012imagenet,vgg}, object detection \cite{he2017mask,dai2016r}, semantic segmentation \cite{badrinarayanan2017segnet,dai2016r} etc.
However, most existing deep learning methods are data-driven and lack of an interpretable way to explain why the networks make a certain prediction.
The interpretability of the deep neural networks or CNNs still needs to be developed so that they can be confidently applied to high-stakes domains such as healthcare, financial services and autonomous driving. 

% There are several existing interpretablity methods which have various motivations. One of the common ideas is to generate a heat map which has the original image's size by calculate the linear combination of activation maps. The activation maps are from the last convolutional layer of Convolutional Neural Networks (CNNs) and have various activations. The heatmap can highlight the area in the image that is most relevant to the CNNs' decisions, including Grad-CAM\cite{gradcam}, Grad-CAM++\cite{gradcamplusplus}, Score-CAM\cite{scorecam}, etc. The difference between this kind of methods is that the weights of the activation maps are different, that is, how to combine the activation maps to better fit the explanation you are looking for.

\begin{figure}[t]
	\centering
	\includegraphics[width=0.78\linewidth]{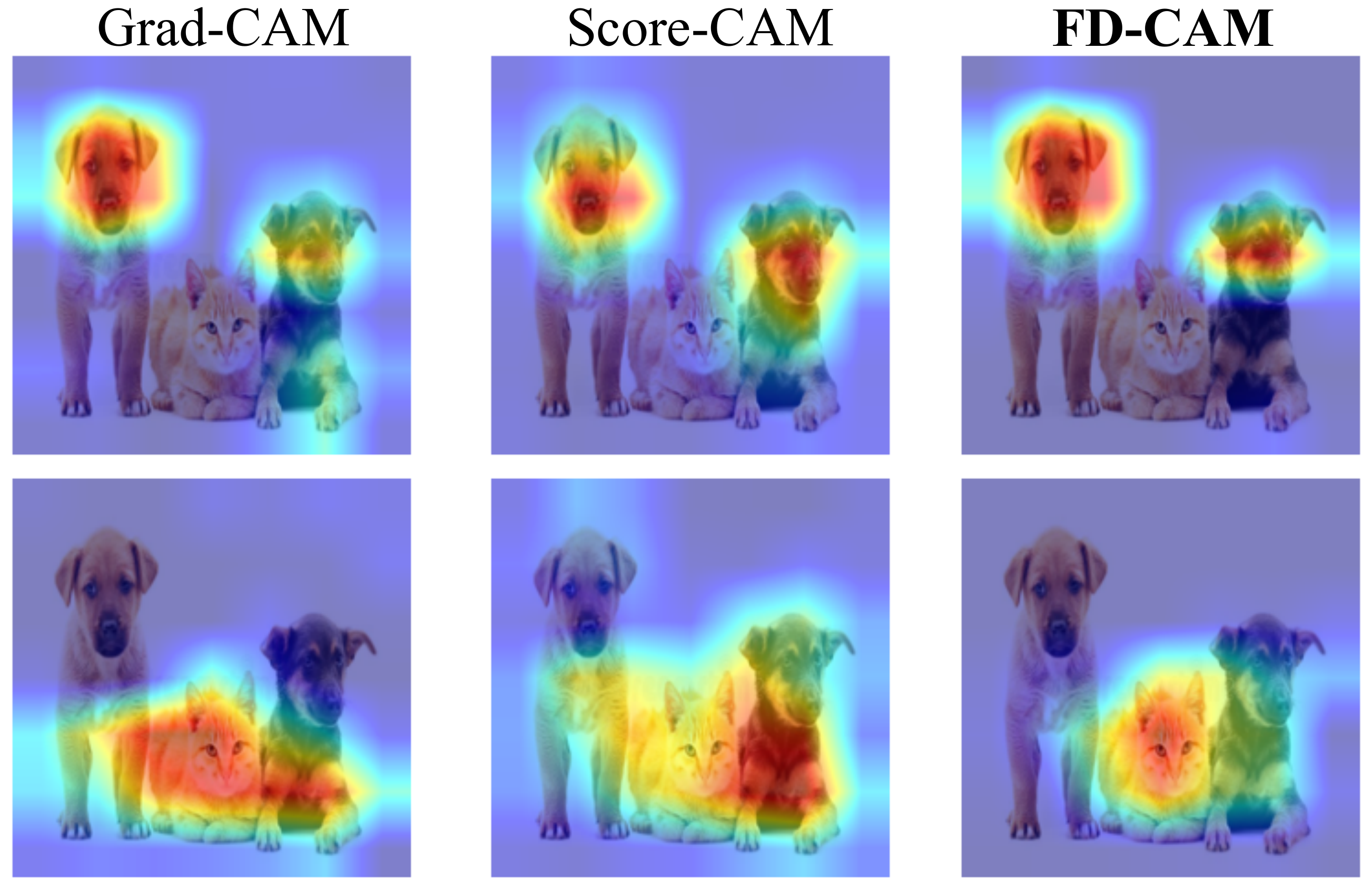}
	\caption{Given an image with multiple targets of different classes, Grad-CAM is more discriminative in explaining the prediction of the CNN model (VGG16) w.r.t different classes and Score-CAM shows more faithful explanation for the multiple targets. Our FD-CAM provides both faithful and discriminative visual explanation of the model. The first row is the CAM visualization w.r.t to the ``Dog'' which has 97.12\% prediction score and the second row is w.r.t to the ``Cat'' whose score is 0.68\%.}
	\label{fig:teaser}
	\vspace{-12pt}
\end{figure}

Visual interpretation or explanation of the internal working mechanism of the deep neural networks, particularly the CNNs, has drawn wide attention and various methods have been proposed.
The early methods focus on interpreting the CNNs via the direct visualization of the gradients in the forms of the saliency map \cite{simonyan2013deep, vgg, zeiler2014visualizing,springenberg2015striving}.
Since the saliency maps generated by the gradient visualization are usually noisy and low quality, following works \cite{sundararajan2017axiomatic,smilkov2017smoothgrad,kapishnikov2019xrai} attempt to improve the sensitivity and smoothness of the results.
Meanwhile, visual explanation of the CNNs by the class activation map (CAM) is another popular technique that can produce more intuitive and high quality results than the gradient visualization.

CAM aims to compute a weighted linear combination of the activation maps in the target convolutional layer of a CNN model and the output is usually a heat map corresponding to the input image.
Typically, the CAM visualization is often used in explaining the classification models so that one can easily get a sense of why the network has made a certain prediction.
The key of CAM-based explanation is to compute effective weights to combine the activation maps w.r.t to the specified class.
And the generated CAM is expected to be faithful and discriminative for explaining the network results.

The original CAM formulation proposed by Zhou et al. \cite{zhou2016learning} requires a retraining step to compute the weights of the activation maps in the last convolutional layer.
Later, various CAM weighting schemes that do not need to modify the network architecture or retraining have been proposed and they can mainly be categorized into gradient and score based methods.
The gradient-based methods compute the gradients of the class prediction score w.r.t each activation map and use the channel-wise global average pooled gradients as the weights.
The gradient-based CAM such as Grad-CAM \cite{gradcam} is discriminative for different class, but may not work faithfully for multiple targets due to the global average pooling operation (see Figure \ref{fig:teaser} left column).
On the other hand, the score-based methods compute the weights by perturbing the input image or feature maps  in the target layer and measuring the changes of the classification scores.
While the score-based CAM such as Score-CAM \cite{scorecam} can improve the faithfulness of the explanation for multiple targets, it is inferior in the discriminability for different class.
The reason is for the class that has a low prediction score in the image (e.g., the ``Cat'' class in Figure \ref{fig:teaser}), its score change may also be insignificant when the perturbation is applied.
Hence, the score-based weights may not be discriminative in this case.

% A representative method based on perturbation is Score-CAM\cite{scorecam}. Score-CAM uses all the activation maps to obtain the pertubation of the original image, and put them back into the network to get the corresponding target class scores as the weights. Score-CAM is to perturb the input of the network therefore is expert in positioning the object, but because the pertubation is not about the internal network, it may be contrary to the purpose of explaining the internal structure of the network, and it takes a long time.

To improve both the faithfulness and discriminability of the CAM-based CNN visual explanation, we propose a novel activation map weighting scheme, named FD-CAM, by combining the merits of the gradient and score based CAM methods.
First, we improve the faithfulness and discriminability of the score-based weights by performing a grouped channel switching operation.
Specifically, for each channel (feature map) in the target layer, we compute its similarity group and switch the group of channels on or off simultaneously to compute changes in the class prediction score as the weights.
% Then, the channels in the same group are switched on and off to compute changes in the class prediction score.
% Finally, we combine the grouped channel switching (GCS) based score with the conventional gradient based weights to compute the final weights for the activation maps.
Then, we combine the improved score-based weights with the conventional gradient-based weights from Grad-CAM so that the discriminability of the final FD-CAM can be further improved. 
We perform extensive comparisons with the state-of-the-art CAM algorithms.
The quantitative and qualitative results show our FD-CAM can produce more faithful and more discriminative visual explanations of the CNNs.
We also conduct experiments to verify the effectiveness of the proposed grouped channel switching and weight combination scheme on improving the results.

% We suggest combining the channel group switching (GCS) based score with the conventional gradient based weights to see if the noise of the gradient weight can be overcome and the class discrimative is guaranteed. Rather than perturbing the input outside the network, we adopt perturbing channel groups of the target convolutional layer in the internal network as the weights based on perturbation. 

% In this paper, we propose FD-CAM, using the combination of gradients with perturbation-based scores as weights. In order to calculate the perturbation-based scores more accurately, we compute the cosine similarity between one piece of feature map and all others. According to the similarity matrix, all channels can be joint with other channels to form groups, when an image to be explained is put into the network, the activations pass through the last convolutional layer, all channel groups will be switched off and on in turn, and the weight of all activation maps in corresponding group is obtained from the target class score. The switch scores and the gradients will be normalized and then combined in form of product. We summarize our contributions as follows:
In summary, our contributions are as follows:
\begin{itemize}
\item We propose FD-CAM, a novel CAM weight scheme which combines the gradient and score based weights to improve the faithfulness and discriminability of visual explanation for CNNs.
\item We introduce the grouped channel switching which perturbs groups of channels simultaneously to obtain more faithful and more discriminative score-based weights.
\item We conduct extensive quantitative and qualitative comparisons with the state-of-the-art CAM algorithms and the results show our FD-CAM can achieve superior performance in explaining the prediction of CNNs.
\end{itemize}
% (1) We introduce a comprehensive CAM-based visual explanation method named FD-CAM, which combine perturbation-based scores and gradients, thus imporving the positioning accuracy and its class discrimative.

% (2) We evaluate the visualization and location capability of FD-CAM with several state of the art CAM-based methods and find FD-CAM perform better. We also test the fairness of generated saliency maps of FD-CAM in form of Deletion and Insertion curve metrics, and do some extensive quantitative experiments to validate FD-CAM can promise the explantions' class discrimative.

% (3) We also perform some ablation study to explore effect of parameters on the FD-CAM, and we provide a mathematical motivation for FD-CAM. Finally, we introduce the effective of its applications as a model validation tool.  

%% file: latex/2_related_work.tex
\begin{figure*}[htbp]
	\centering
	\includegraphics[width=0.85\linewidth]{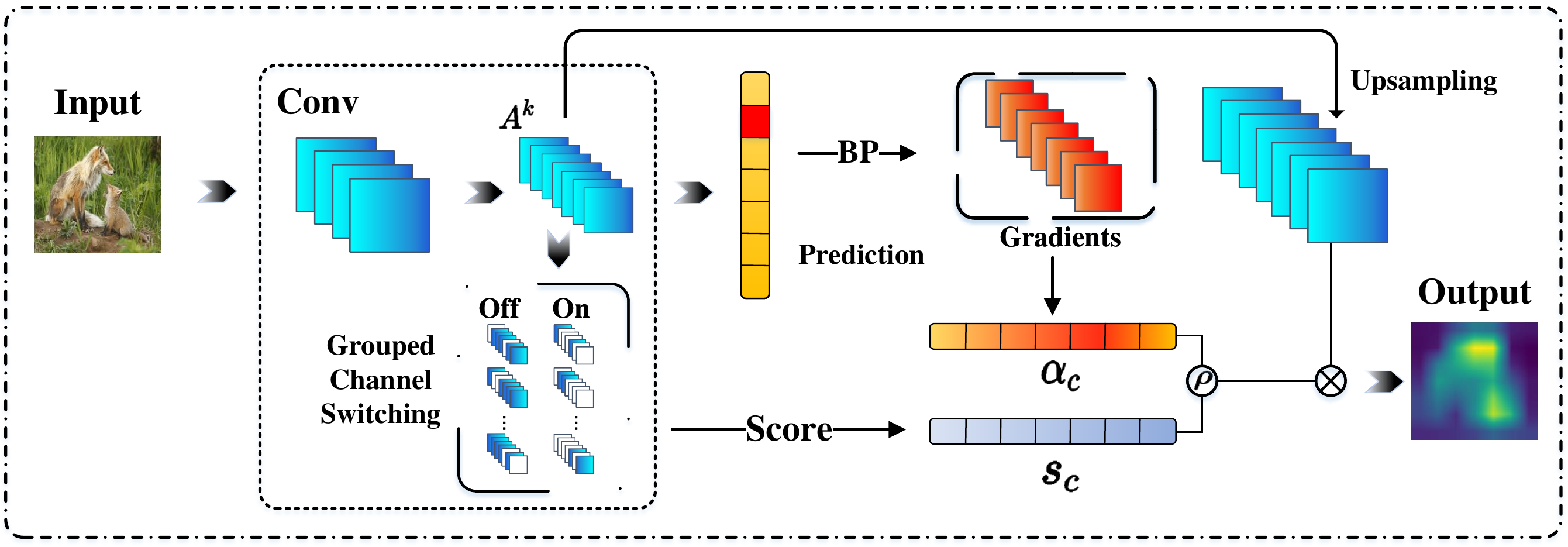}
	\caption{The pipeline of FD-CAM. Given an input image and a CNN model (e.g., VGG16), we use the conventional gradient-based method to compute the gradient-based weights $\alpha_c$.
% 	the gradients w.r.t the activation maps $A^k$ in the target convolutional layer are computed by backpropagation, while the gradient-based weights $\alpha_c$ are obtained by global average pooling.
	Then, a novel grouped channel switching is employed to compute the prediction score change as the weights $s_c$.
	Finally, we combine $\alpha_c$ and $s_c$ to obtain the final weights which can be used to linearly combine the upsampled activation maps to produce the visual explanation in the heat map form.
% 	Activations and gradients are first extracted through forward-passing and backpropagation. Then the similarity matrix of activations is used to split them into several groups. Those groups are switched to get the perturbed scores, combining the gradients to form overall weights. Finally, the result can be generated by a linear combination of overall weights and upsampled activation maps.
	}
	\label{pipeline}
	\vspace{-6pt}
\end{figure*}

\section{Related Work}
In this section, we first summarize a generalized CAM formulation.
Then, we briefly survey on the existing gradient and score based CAM methods with the focus on how they compute the activation map weights and their performance in the visual explanation of CNN classification models.

% \noindent\textbf{CAM revisited.}
% Assuming $X$ is an input image, $f$ is a CNN model, then $f_c(X)$ represents the score generated by CNN $f$  of class $c$. First, we compute the gradients of score $f_c(X)$ with respect to the feature map $A^k \in R^{h \times w}$, then calculate the global average of the gradient to obtain the weight of each feature map,
% \begin{equation}
%     \alpha_{c}^{k}=\frac{1}{h \times w} \sum_{i} \sum_{j} \frac{\partial f_{c}\left(X_{0}\right)}{\partial A_{i j}^{k}\left(X_{0}\right)}
% \end{equation}
% where $h, w$ are the length and width of feature map $A_k$ respectively.

\textbf{Generalized CAM formulation.} Assume $f(X)$ is a CNN model which takes an input image $X$ and predicts the probabilities or scores of different classes.
For the target convolutional layer in $f$ which contains $K$ feature maps or channels of spatial size $h \times w$, define $A^k \in R^{h \times w}$ as the $k$-th feature map.
Normally, the generalized formulation of the class activation map of the CNN model $f$ w.r.t class $c$ can be defined as:
\begin{equation}
\mathcal{L}_{\text {Gen-CAM }}^{c} = \Phi\left(\sum_{k \in K} \omega^k_c A^k \right),
\end{equation}
where $\Phi(\cdot)$ is an activation function, $\omega^k_c$ are the weights to combine the activation maps $A^k$.

In the original CAM \cite{zhou2016learning}, $\Phi(\cdot)$ is defined as an identity function and $\omega^k_c$ is a set of retrained weights for classifying the features obtained by applying global average pooling on each channel of the last convolutional layer.
Inspired by \cite{zhou2016learning}, numerous CAM variants have been proposed, in which $\Phi(\cdot)$ is usually defined as the ReLU function to focus on the features with positive influence.
For $\omega^k_c$, various schemes including the gradient and score based methods have been investigated to find effective weights which can produce more discriminative CAMs with faithful  visual explanation.

\textbf{Gradient-based CAM.} 
In contrast to \cite{zhou2016learning} which requires to retrain a modified model to obtain the weights, Selvaraju et al. propose Grad-CAM \cite{gradcam}, the first gradient-based weighting scheme, to define $\omega^k_c$ as the channel-wise global average pooled gradients $\alpha^k_c$ of the model prediction $f_c(X)$ w.r.t the activation map $A^k(X)$ for class $c$:
\begin{equation}
    \omega^k_c=\alpha_{c}^{k}=\frac{1}{h \times w} \sum_{i} \sum_{j} \frac{\partial f_{c}\left(X\right)}{\partial A_{i j}^{k}\left(X\right)}.
\end{equation}
Using the gradients as weights, Grad-CAM can be applied to different CNN involved tasks beyond the image classification in \cite{zhou2016learning}, such as image captioning \cite{johnson2016densecap} and visual question answering \cite{antol2015vqa}.
In addition, Grad-CAM computes per-pixel gradients of the activation map w.r.t to the given class.
Therefore, it is discriminative to different classes.
On the other hand, because of the global average pooling, Grad-CAM may discard the spatial information which can make the CAM-based localization not faithful to multiple object targets.

To solve this issue, Chattopadhyay et al. propose Grad-CAM++ \cite{gradcamplusplus} which employs weighted average of the positive gradients as $\alpha_c^k$, while the weights of the gradients are calculated by the higher-order derivatives of $f_c(X)$ w.r.t $A^k(X)$.
With the weighted combination of the gradients, Grad-CAM++ has shown more faithful results on explaining images with multiple targets than Grad-CAM. Additionally, CAMERAS\cite{cameras} performs multi-scale accumulation and fusion of the activation maps and backpropagates gradients to compute high-fidelity saliency maps.
In our FD-CAM, we aim to improve the faithfulness of Grad-CAM while keeping its discriminability.
Different from using the higher-order derivatives which may be unstable to compute, we combine the score-based weights with the gradient-based weights so that the faithfulness of the CAM can be improved.

\textbf{Score-based CAM.} 
\textit{Perturbation} is a way to generate the variants of the input or the model so that the change of the model's prediction score can be used as an indicator for the importance of the perturbation operation. 
For example, in RISE \cite{RISE}, random binary masks are sampled to generate the perturbed masked input images and the output prediction scores are used as importance weights to combine the sampled masks to get the visual explanation of the black-box model.
In the score-based CAM methods, $\omega^k_c$ is defined as the changes of the classification scores caused by perturbing the input image \cite{fong2017interpretable, fong2019understanding, scorecam,groupcam} or feature maps \cite{desai2020ablation} in the target layer, i.e., 
\begin{equation}
\omega^k_c = s_{c}^{k},
\end{equation}
where $s_{c}^{k}$ is classification score change w.r.t class $c$ and it will be used as the weight for activation map $A^k$.
Instead of performing backpropagation to compute the gradients in the gradient-based method, $s_{c}^{k}$ is calculated by applying forward propagation and comparing the original classification score with the perturbation-induced score.

Score-CAM \cite{scorecam} is one representative score-based method which performs the perturbation by multiplying the input image with each activation map in the target convolutional layer and then combines the activation maps using the change of the model's confidence score as the weights.
Meanwhile, Desai et al. propose Ablation-CAM \cite{desai2020ablation} that perturbs the feature maps in the target layer.
Each feature map or unit $A^k$ is disabled or switched off in turn to compute score changes as the weight for this unit.
LIFT-CAM\cite{jung2021towards} formulates the explanation model as a linear function of binary variables denoting the existence of the associated activation maps, while the weights of each activation map is derived from the SHAP\cite{lundberg2017unified} values. 
These methods compute the weight $\omega^k$ based on only manipulating the activation map $A^k$.
In contrast, we perform grouped channel (feature map) perturbation and introduce the group switch-off and switch-on operation to compute the score change.
To further improve the discriminability of the score-based methods, we combine the scores obtained by the grouped switching and the gradient-based weights to compute the final weights for FD-CAM.

%% file: latex/3_method.tex
\section{Method}

In this section, we first discuss the motivation of channel grouping and introduce how to form the groups based on cosine similarity between different channels.
Then, we apply the grouped channel switching operation to compute the change of class classification scores as the improved score-based weights.
%which will be used to adjust the existing gradient based weights.
Finally, we propose a novel CAM weighting scheme by combining the gradient-based weights with score-based weights.
Figure \ref{pipeline} shows the pipeline of our FD-CAM.

% In this section, we describe the FD-CAM algorithm, and then we show its pipeline in detail in Fig \ref{pipeline} and the high-level steps in Algorithm \ref{FD-CAM}. After that, the motivation behind FD-CAM will be explained.

% After we get $\alpha_c^k$, normalize $\alpha_c^k$ as
% \begin{equation}
% \alpha_{c}
% =\frac{\alpha_c-\min \left(\alpha_c\right)}{\max \left(\alpha_c\right)-\min \left(\alpha_c\right)}
% \end{equation}

% In order to make the weights keep class discrimative, we map $\alpha_{c}$ to $[-1,1$],
% \begin{equation}
% \alpha_{c}^{'}= 2\alpha_{c}-1
% \end{equation}
% % \vspace{6pt}

% \noindent\textbf{Channel grouping and switching.}
\subsection{Channel Grouping}
\label{grouping}
\begin{figure}[t]
	\centering
	\includegraphics[width=0.75\linewidth]{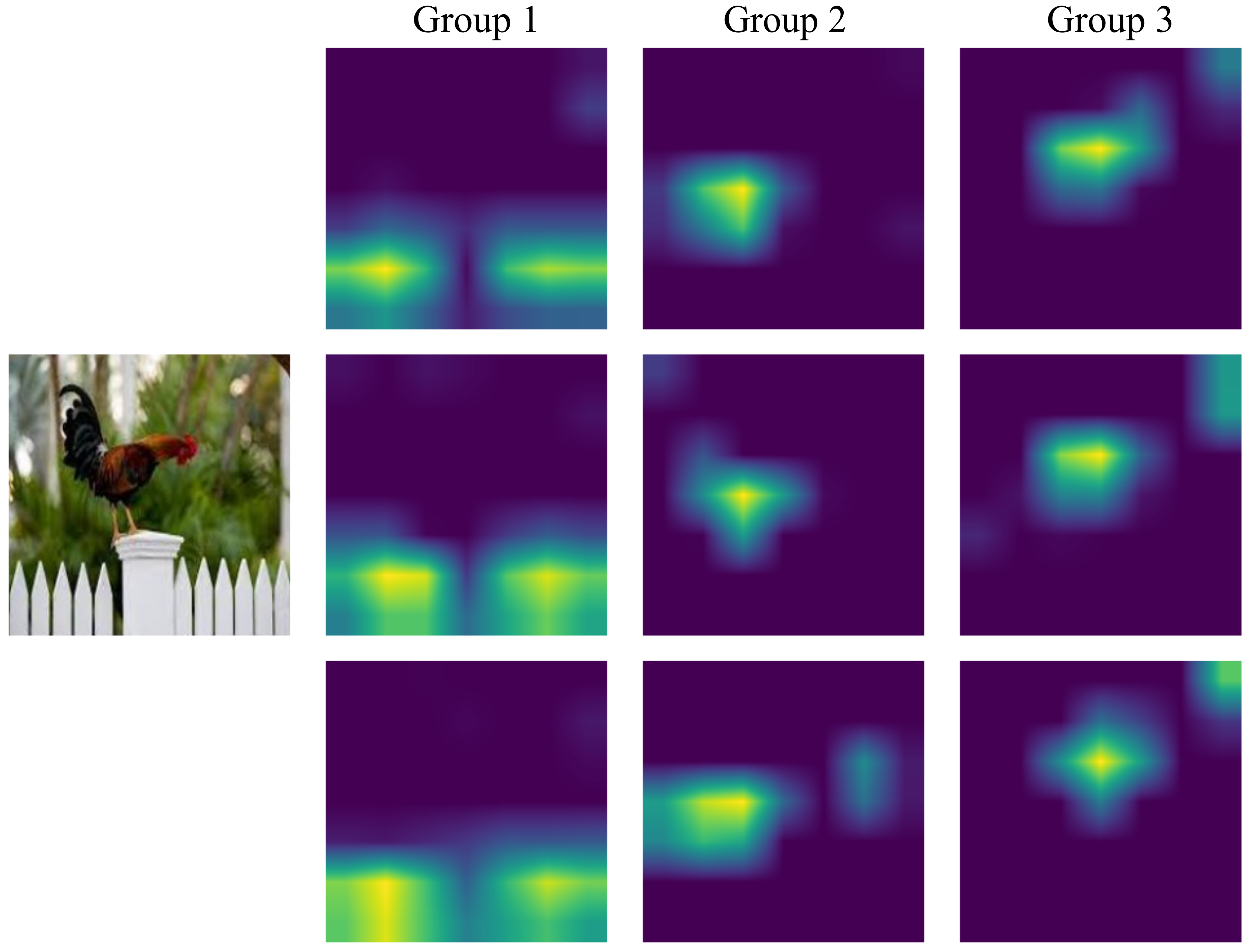}
	\caption{Visualization of features in three channels (first row) of the target convolutional layer and the groups formed by similar features in other channels (second and third row). From each group (column), it can be observed there are indeed very similar channels existing in the target layer and they correspond to the similar regions in the image. }
	\label{fig:group_vis}
	\vspace{-6pt}
\end{figure}

\textbf{Motivation}. Inspired by the Ablation-CAM \cite{desai2020ablation} which computes the score-based weights by switching off each individual channel and measuring the classification score changes, we perform grouped channel switching operation to find the importance of each channel in a more discriminative manner.
The reason for using the grouped switching is, in a particular convolutional layer, the features in different channels may actually have high similarities (see Figure \ref{fig:group_vis}).
If only one channel is switched off, the final classification score may not change much, since other channels may still have the similar features activated.
Moreover, performing the grouped switching can be regarded as considering more context when computing the importance for each feature map $A^k$.
Hence, the score changes may better represent the importance of the interested channel if all of its similar channels are switched on or off together.

\textbf{Channel similarity.} Specifically, to determine the weight for each channel $A^k$, we simultaneously switch on or off a group of channels which have high similarities to $A^k$, and compute the score changes as $s_c^k$.
The similarity between channel $A^k$ and another channel $A^l$ is defined by the cosine similarity:
\begin{equation}
cos(A^k,A^l) = \frac{\mathbf{v}^k \cdot \mathbf{v}^l}{\|\mathbf{v}^k\| \cdot \|\mathbf{v}^l\|},
\end{equation}
where $\mathbf{v}^k$ and $\mathbf{v}^l$ are the vectors obtained by flattening the matrices $A^k$ and $A^l$.
Then, a cosine similarity matrix $M$ is computed to store the similarities between each pair of channels in the target layer.

Next, the similarity group of $A^k$ is defined as
\begin{equation}
    G(A^k) = \{A^l | cos(A^k, A^l) > \tau_\theta^k, l, k \in K\}.
\end{equation}
Here, $\tau_\theta^k$ is a similarity threshold determined by the $\theta$-th percentile of the high-to-low sorted similarities for channels in $G(A^k)$.
Empirically, we use $\theta=5$ to set top 5\% similar channels to be in the same group.
Note that, for each channel, we find its own group for the score change computation.

\subsection{Grouped Channel Switching}
Similar to \cite{desai2020ablation}, for each $A^k$, we first switch off all the channels in its similarity group $G(A^k)$ and compute the score change as the \textit{switch-off score}:
\begin{equation}
    s_c^{k-} = f_c(X)-f_{c}^{k-}(X),
\end{equation}
where $f_c(X)$ is the original classification score of input $X$ w.r.t class $c$ and $f_{c}^{k-}(X)$ is the new classification score from the modified model with channels in $G(A^k)$ switched off.

In addition to switching off or dropping the group of similar features for evaluating the importance of $A^k$, we also propose a new grouped channel \textit{switch-on score}:
\begin{equation}
    s_c^{k+} = f_{c}^{k+}(X),
\end{equation}
where $f_{c}^{k+}(X)$ is the classification score when only the channels in $G(A^k)$ are switched on and other channels are switched off.

Finally, we define our grouped channel switching based score as
\begin{equation}
    s_c^{k} = \frac{1}{2}(s_c^{k-} + s_c^{k+}).
\end{equation}
By combining $s_c^{k-}$ and $s_c^{k+}$, we can evaluate the importance of each channel from two perspectives: switch off and on to deactivate and activate the influence, respectively.
Moreover, the grouped channel switching treats the feature maps in a larger channel-wise context, while the influence of one channel is associated with all of its similar channels.
Conceptually, when our channel group is set to be the interested channel itself, our grouped channel switching is reduced to the Ablation-CAM, while the only difference is how we compute the score $s_c^{k}$.

\subsection{Combination of Gradient and Score based Weights}
\label{FD-CAM}

In our FD-CAM, we combine the gradient-based weights $\alpha_{c}^k$ and score-based weights $s_c^k$ to define $\omega_c^k$ so that the merits of both methods can be highlighted:
\begin{equation}
\omega^k_c = \rho (\alpha_{c}^k, s_{c}^{k}),
\end{equation}
where $\rho(\cdot, \cdot)$ is a function that combines $\alpha_{c}^k$ and $s_{c}^{k}$.
%and we will introduce its definition in Section \ref{FD-CAM}.

Since the original gradient-based and score-based weights are in different scales, to combine them properly, we apply the standard min-max normalization to $\alpha_{c}^k$ and $s_{c}^{k}$ and obtain $\hat{\alpha}_{c}^k$ and $\hat{s}_{c}^{k}$ for which the values are linearly scaled to $[0,1]$.
% Next, there are multiple options to define the function $\rho(\cdot, \cdot)$ such as a linear or an exponential based function which 
Next, we define
\begin{equation}
\label{eq:combine}
\rho (\alpha_{c}^k, s_{c}^{k}) = \hat{\alpha}_{c}^k e^{\hat{s}_{c}^{k}} - b,
\end{equation}
where $b$ is a bias parameter (empirically set to 0.5) to allow negative weights for combining the activation maps.
Generally, there are multiple options to combine the two different types of weights.
In our formulation, we treat the $e^{\hat{s}_{c}^{k}}$ as a special scaling weight for $\hat{\alpha}_{c}^k$, while the influence of $\hat{s}_{c}^{k}$ is incorporated in an exponential manner.
In Section \ref{sec:ablation}, we show our proposed formulation outperforms other options by comparing the results of using different parameters $b$ and different weight combination methods.

Finally, like other CAM methods, we define our FD-CAM for target convolutional layer $\{A^k| k \in K \}$ w.r.t class $c$ as 
\begin{equation}
\mathcal{L}_{\text {FD-CAM }}^{c} = \text{ReLU}\left(\sum_{k \in K} \rho (\alpha_{c}^k, s_{c}^{k}) A^k \right).
\end{equation}

% \rui{Need to update... 
% % Since the calculation of $\omega_c$ is based on target class score, $\omega_c$ has a high degree of fidelity with the regions that relate to the class $c$ in the input image $X$. But $\omega_c$ is not discriminative to other classes with low probability in the network result, which is a severe defect as an explanation method. Therefore, it needs to be combined with $\alpha_{c}^{'}$, and then we make Hadamard product of $\omega_c$ and $\alpha_{c}$ to enhance the accuracy of positioning the relevant area.
% % \begin{equation}
% % \beta_c = \alpha_{c}^{'} + \sum_{i=1} \frac{2}{i!\cdot s(i)} \alpha_c \otimes (\omega_c)^i
% % \label{wpp}
% % \end{equation}
% % where $(\omega_c)^i$ is element-wise exponentiation and $s(i)$ is a positive proportional function of $i$.

% Like all the CAM-based techniques, explanation is the weighted linear combination as:
% \begin{equation}
% \mathcal{L}_{\text {FD-CAM }}^{c} = \text{ReLU}\left(\sum_k \beta^k_c A^k_{H\times W} \right)
% \end{equation}
% where $A^k_{H\times W}$ is the result of bilinear interpolation of $A_k$, $H,W$ are the height and width of the original image.
% }

%% file: latex/4_experiments.tex
\section{Experiment}

\begin{figure*}[htbp]
	\centering
	\includegraphics[width=0.9\linewidth]{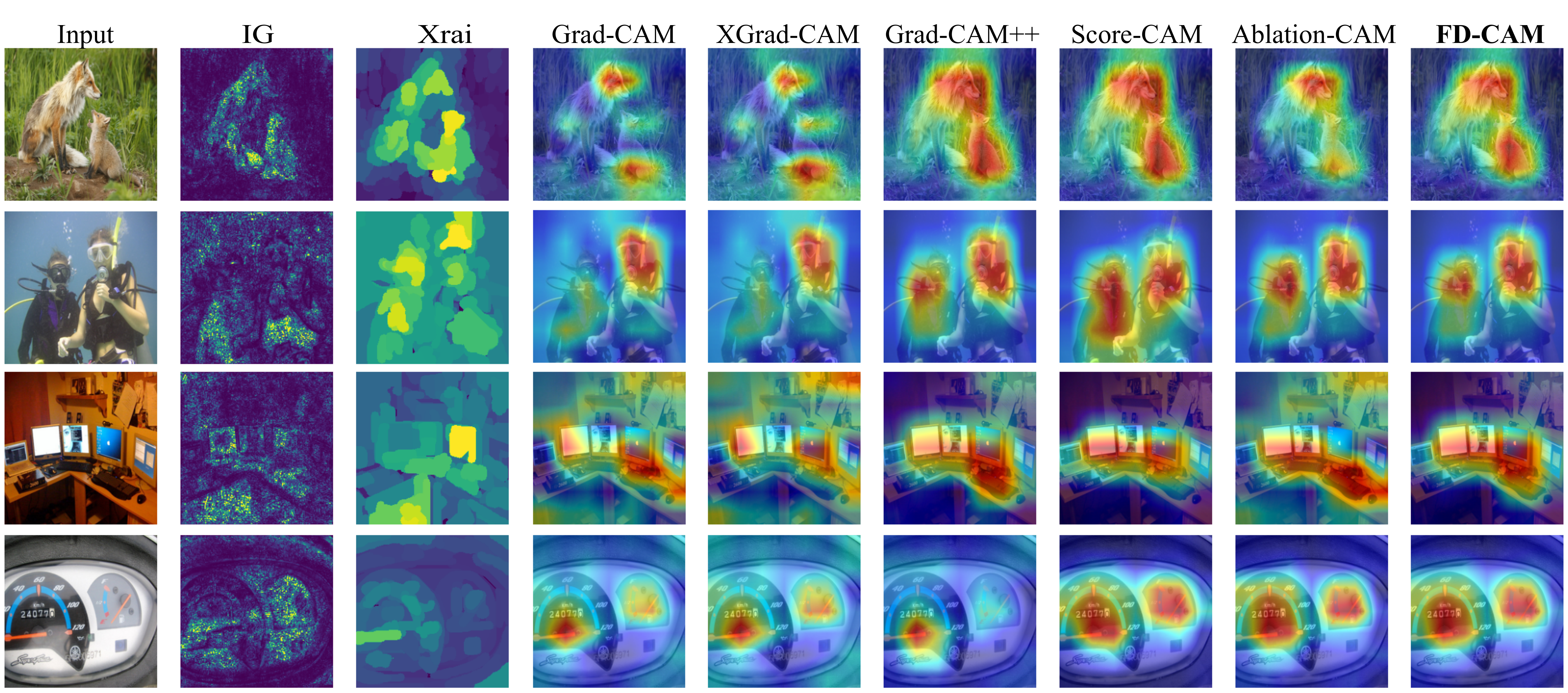}
	\caption{
% 	Some visualization explanation generated by some SOTA  saliency methods including Integrated Gradients,  XRAI, Grad-CAM, XGrad-CAM, Grad-CAM++, Score-CAM, Ablation-CAM and FD-CAM for "red fox", "scuba diver", "desktop computer" and "odometer". For an input image containing several objects of the same class, FD-CAM can completely and accurately mark the objects with less noise.
Qualitative comparisons of FD-CAM with other SOTA CAM results on visual explanation of images w.r.t specified class (from top to bottom, the classes of interest are ``Red fox'', ``Scuba diver'', ``Desktop computer'', ``Odometer'').
Our FD-CAM produces more faithful results (i.e., more accurately highlight all related regions) than other methods.
	}
	\label{visulization}
\vspace{-10pt}
\end{figure*}

% In this section, we propose several requirements that need to be met for good explanations. They are the localization ability, the faithfulness of the explanation to the original result and the ability of distinguishing different classes of objects of explanation. After that, we carry out experiments from these aspects to compare the proposed FD-CAM with other popular saliency map methods. The experiment conclusions show that FD-CAM is a good interpretable method, and can take the lead among the current CAM-based methods when several interpretability capabilities are considered comprehensively.

In this section, we first introduce the implementation details of FD-CAM. Then, we compare our results with SOTA methods on visual explanation of CNNs. In addition, we perform qualitative and quantitative evaluation on the faithfulness and discriminability of our method.
Finally, we conduct ablation studies to show the effectiveness on the grouped channel switching and the weight combination scheme.

% Our code is available at \href{ https://anonymous.4open.science/r/FD-CAM}{https://anonymous.4open.science/r/FD-CAM}.

% \subsection{Experiment Setting}
\textbf{Implementation details.}
In the following, except for the quantitative evaluation on discriminability,  we use 2,000 images randomly selected from the ILSVRC2015 validation set \cite{ILSVRC15} for the qualitative and quantitative evaluation.
For the discriminability evaluation, we experiment on the VOC2007 validation set\cite{pascal-voc-2007} which contains more images with multiple categories.
During the pre-processing, all the images are resized to $224\times 224 \times 3$.
For the CNN model to be explained, we choose the pre-trained VGG16 \cite{vgg} from the PyTorch model zoo.
We implement FD-CAM in PyTorch and conduct all experiments on a desktop with 1 NVIDIA TITAN RTX GPU.
For other methods, we adopt their open source implementation and test each method on our datasets.
% The experiment dataset in this section is chose from a commonly used dataset in computer vision task. We choose the first 2000 images in the ILSVRC validation set\cite{ILSVRC15} as experiment dataset. In the stage of preprocessing, all the images are resized to $3\times 224\times 224$, and then converted to tensor and normalized to [0,1]. In all experiments, the pretrained torchvision model VGG16\cite{vgg} are chose to be the model to be explained.

\begin{table}[]
\centering
\footnotesize
\caption{The AUC values of insertion (higher is better), deletion (lower is better) and the overall metric (insertion - deletion, higher is better).
The best values are in \textbf{bold}, and second best are in \textit{italic}.
}
\label{delinsexpriment}
\begin{tabular}{@{}lcccc@{}}
\hline
Methods              & Insertion $\uparrow$     & Deletion $\downarrow$        & Overall $\uparrow$        \\ 
\hline
Grad-CAM\cite{gradcam}               & 0.5357    & 0.1117          & 0.4240          \\
Grad-CAM++ \cite{gradcamplusplus}    & 0.5321          & 0.1088          & 0.4233          \\
XGrad-CAM \cite{Xgradcam}            & 0.5464          & 0.1072          & 0.4392          \\
Score-CAM \cite{scorecam}            & 0.5422          & 0.1059          & 0.4363          \\
Ablation-CAM \cite{desai2020ablation}  & \textit{0.5502}    & 0.1013          & \textit{0.4489}         \\
Layer-CAM  \cite{jiang2021layercam}        & 0.5389    & 0.1021          & 0.4368          \\
Group-CAM \cite{groupcam}         & 0.5397          & \textbf{0.0921}     & 0.4476         \\
FD-CAM           & \textbf{0.5534}          & \textit{0.1001}          & \textbf{0.4533}\\
\hline
\end{tabular}
\vspace{-6pt}
\end{table}
% \begin{table}[]
% \centering
% \footnotesize
% % \setlength{\tabcolsep}{0.03\linewidth}
% \caption{The AUC values of insertion (higher is better), deletion (lower is better) and the overall metric (insertion - deletion, higher is better).
% The best values are in \textbf{bold}, and second best are in \textit{italic}.
% }
% \label{delinsexpriment}
% \begin{tabular}{@{}lccccc@{}}
% \hline
% Methods              & Insertion $\uparrow$     & Deletion $\downarrow$        & Overall $\uparrow$  &Acc(\%) $\uparrow$\\ 
% \hline
% Grad-CAM\cite{gradcam}               & 0.5357    & 0.1117          & 0.4240    & 81.20      \\
% Grad-CAM++ \cite{gradcamplusplus}    & 0.5321          & 0.1088          & 0.4233  & 81.91        \\
% XGrad-CAM \cite{Xgradcam}            & 0.5464          & 0.1072          & 0.4392  & 80.72        \\
% Score-CAM \cite{scorecam}            & 0.5422          & 0.1059          & 0.4363  & 78.46        \\
% Ablation-CAM \cite{desai2020ablation}  & \textit{0.5502}    & 0.1013          & \textit{0.4489}  & 58.19       \\
% Layer-CAM  \cite{jiang2021layercam}        & 0.5389    & 0.1021          & 0.4368     & 81.77     \\
% Group-CAM \cite{groupcam}         & 0.5397          & \textbf{0.0921}     & 0.4476     & -    \\
% FD-CAM           & \textbf{0.5534}          & \textit{0.1001}          & \textbf{0.4533} & \textbf{83.70}\\
% \hline
% \end{tabular}
% \vspace{-6pt}
% \end{table}

\subsection{Comparison with SOTA on Visual Explanation}
% \textbf{Qualitative comparison.}
In Figure \ref{visulization}, we show the heat map visualization of the results for some images in ILSVRC.
We compare the FD-CAM with different types of CAM (or saliency map) methods, including the gradient-based visualization (Integrated Gradients \cite{sundararajan2017axiomatic}, XRAI \cite{ kapishnikov2019xrai}), gradient-based CAM (Grad-CAM \cite{gradcam}, Grad-CAM++ \cite{gradcamplusplus} and  XGrad-CAM \cite{Xgradcam}) and score-based CAM (Score-CAM \cite{scorecam}, Ablation-CAM \cite{desai2020ablation}).
It can be observed the IG and XRAI results are more noisy and less useful comparing to the CAM-based methods.
In Grad-CAM, Grad-CAM++ and XGrad-CAM, the visualization may not highlight all related targets (e.g., the odometer on the right).
Score-CAM and Ablation-CAM produce better results for multiple targets situation, but they may still include less focused regions.
Our FD-CAM can faithfully and more accurately highlight all related regions w.r.t the specified class.

% We randomly choose several images from ILSVRC validation set \cite{ILSVRC15} for some SOTA CAM-based methods and FD-CAM. The visual results can be seen in Fig. \ref{visulization}.
% It can be obviously observed that the emphasis of FD-CAM focused more on the object, compared to its counterparts, is equipped with smaller noise. 
% Specially, for images that contain multiple objects, FD-CAM can also accurately highlight them simultaneously.
% These properties are quite valuable for the visual explanation of images.

\begin{figure}[htbp]
	\centering
	\includegraphics[width=0.95\linewidth]{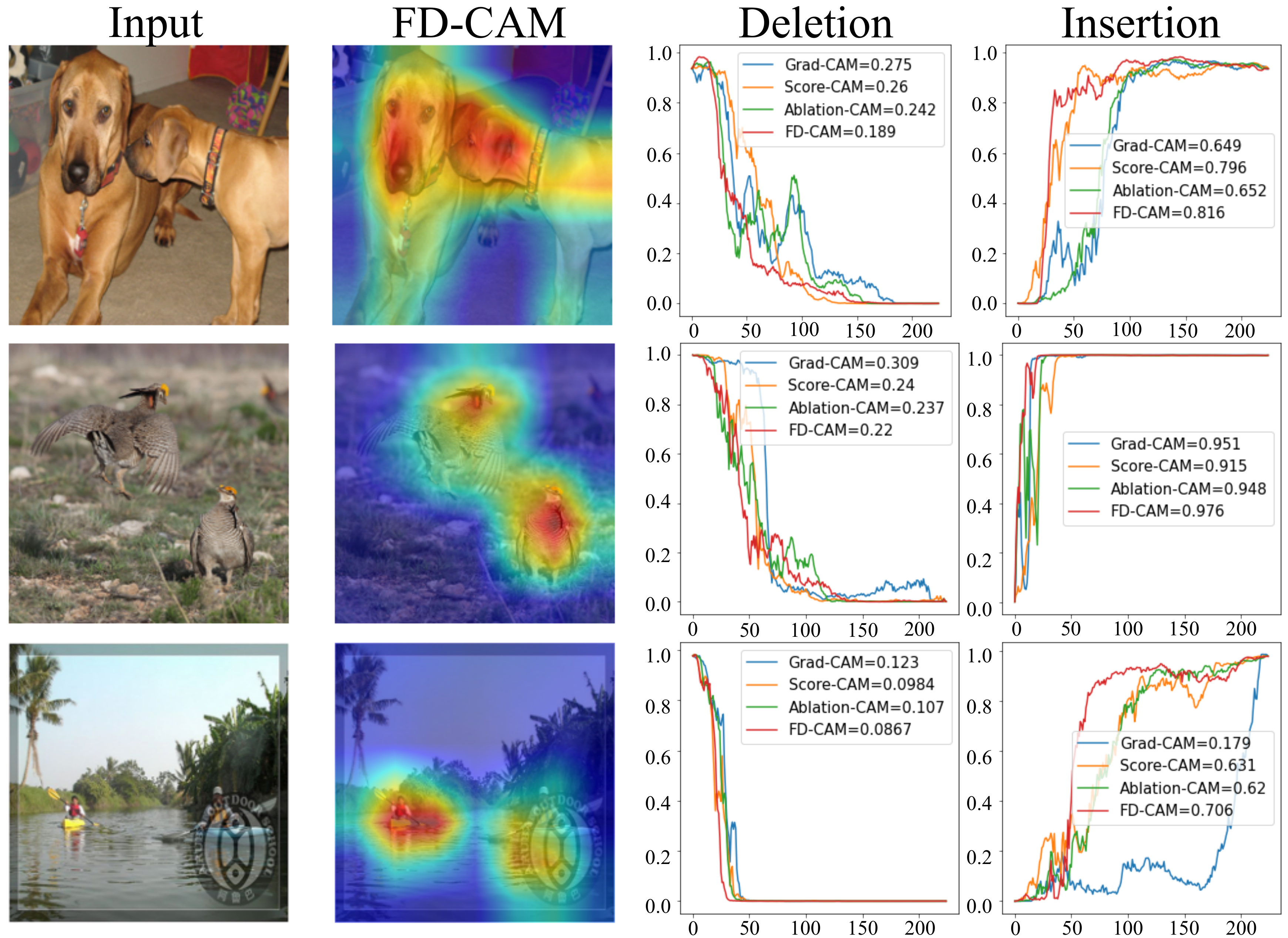}
	\vspace{-2pt}
	\caption{Evaluation on faithfulness by comparing the deletion and insertion curves.
	Our FD-CAM produces steeper curves and better AUC (shown in the legend) than other methods, showing the high-activated regions in FD-CAM are more faithfully related to the model's decision.
	From top to bottom, the interested classes are ``Rhodesian ridgeback'', ``Prairie chicken'' and ``Canoe''.
% 	It's interesting to see FD-CAM can highlight the paddle area on the right of the last image, even if noises such as watermark exist. 
% 	Some CAM-based methods generated saliency maps for representative images including "Rhodesian ridgeback", "prairie chicken" and "canoe" with deletion and insertion curves.  A faithful explanation is expected to drop faster and the AUC will be small in terms of deletion curve, while in increase curve, the curve turn to surge and the AUC will be large.
	}
	\label{curveexample}
	\vspace{-12pt}
\end{figure}

\begin{table*}[!t]
\centering
\small
\setlength{\tabcolsep}{4pt}
\caption{The evaluation on discriminability using the pointing games metric.}
\vspace{-4pt}
\begin{tabular}{@{}cccccccc@{}}
\hline
  & Grad-CAM \cite{gradcam} & Grad-CAM++ \cite{gradcamplusplus}  & XGrad-CAM \cite{Xgradcam} &  Layer-CAM \cite{jiang2021layercam}  & Score-CAM \cite{scorecam} & Ablation-CAM \cite{desai2020ablation} & FD-CAM \\
\hline
Acc(\%) $\uparrow$ & 81.20 & 81.91 & 80.72 & 81.77 & 78.46 & 58.19 & \textbf{83.70} \\
\hline
\end{tabular}
\label{acc}
\vspace{-10pt}
\end{table*}  

\subsection{Evaluation on Faithfulness}
One important characteristic to evaluate the visual explanation technique is the faithfulness, which measures how the results can accurately highlight the regions related to the model's decision.
To quantitatively evaluate the faithfulness, we adopt the metric proposed in \cite{RISE}:
the area under the deletion and insertion curve (AUC).
The deletion curve shows the decrease of class prediction probability when gradually deleting the high-activated regions from the input image, while the insertion curve shows the increase of probability when inserting the regions to a zero-valued image or blurred version of input.
A faithful explanation is expected to start with a sharp drop in predicted probability at the start of the deletion curve, which corresponds to a smaller AUC.
At the start of the insertion curve, the predicted probability is expected to increase quickly which can result in a larger AUC. 

% Undoubtedly, faithfullness is quite an important property for an explanation.
% If the saliency map can not reflect precisely the results of the original network, then it will easily lead to misunderstandings of the original network. 
% When assessing the faithfullness, we adopt the metric proposed in  \cite{RISE}. 
% The area under deleting and inserting curve (AUC) represents decrease and increase of probability when gradually deleting and inserting feature areas from the original input to the reference input (e.g. constant-value image or blurred image).
% A faithful explanation is expected to start with a sharp drop in predicted probability at the start of deletion, corresponding to a smaller AUC, while at the start of insertion, the predicted probability surge violently, and hence a larger AUC. 
% For example, the form of the deletion or insertion curve is shown in Fig \ref{del ins example}.

% \begin{figure}[htbp]
% 	\includegraphics[width=1.0\linewidth]{fig/delinsexample.pdf}
% 	\caption{(b) and (c) show the change curves of the prediction results when pixels are gradually inserted and deleted in the order of pixel importance in the saliency map of input(a).}
% 	\label{del ins example}
% \end{figure}

We evaluate the faithfulness of the results from several CAMs and FD-CAM by deleting or inserting regions with a rate of 3.6\% of the original image's area at each step.
The quantitative results are shown in Table \ref{delinsexpriment}.
We also show some CAM visualizations along with their deletion and insertion curves in Figure \ref{curveexample}.
It can be observed FD-CAM can generally outperform other methods on faithfulness.

\begin{figure}[t]
\centering
\vspace{-6pt}
	\includegraphics[width=0.85\linewidth]{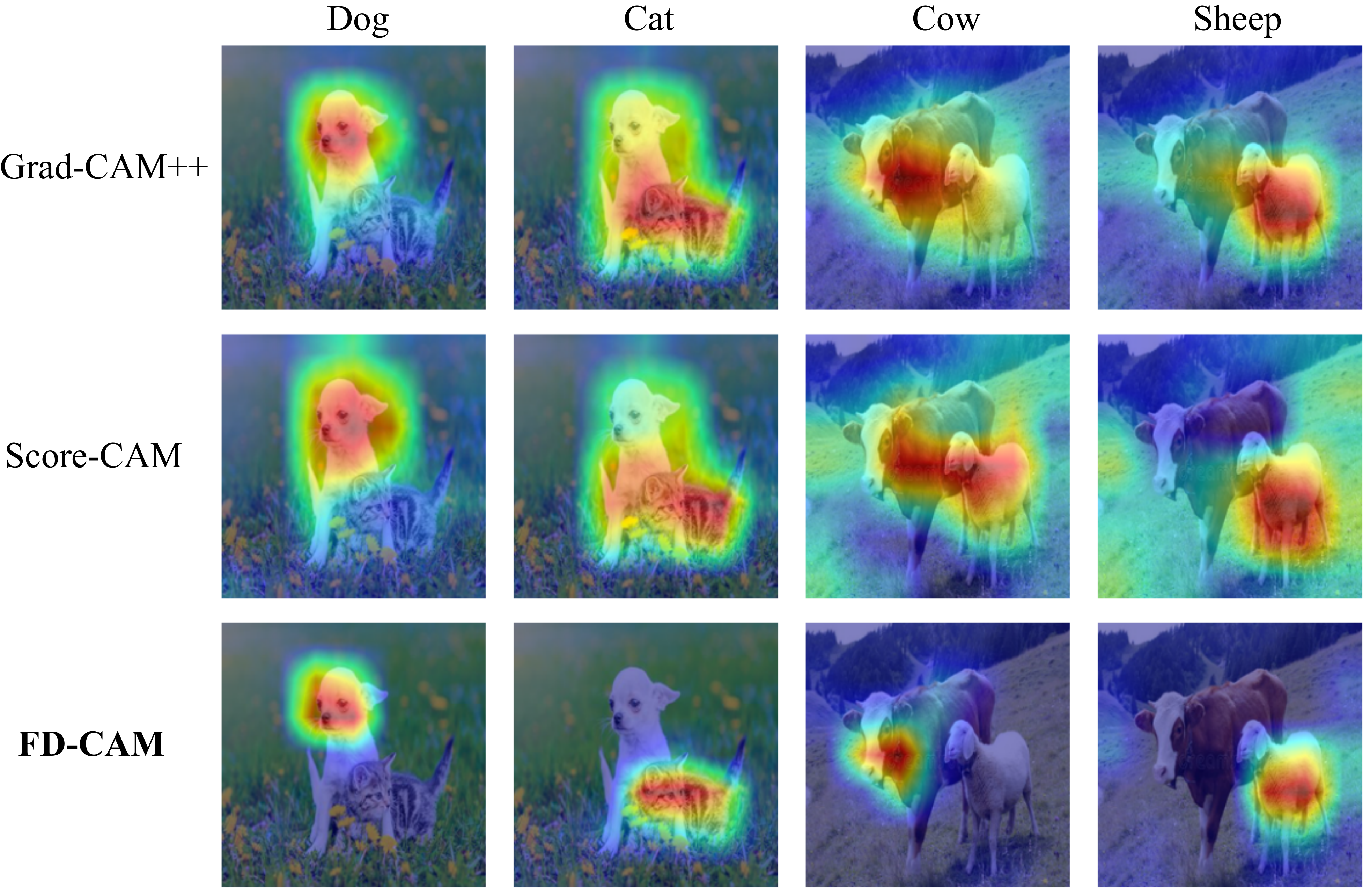}
	\vspace{-2pt}
	\caption{Qualitative evaluation on class discriminability. FD-CAM can produce more discriminative results for different classes in the same image.}
	\label{dogcat}
	\vspace{-12pt}
\end{figure}

\subsection{Evaluation on Discriminability}
Discriminability on different classes is another critical property for evaluating the performance of the visual explanation methods.
Given an input image, it usually contains multiple classes with different prediction probabilities.
For the non-dominant class in the image, even its probability may be low, an effective method is still expected to correctly visualize the regions corresponding to the model's decision.
Figure \ref{dogcat} shows the qualitative evaluation on the class discriminability for FD-CAM.
It is clear that FD-CAM outperforms other gradient and score based methods in distinguishing different classes when explaining the model's decision.

For quantitative evaluation, we employ the localization-based metric pointing game \cite{zhang2018top} to compute the point-based localization accuracy for different classes as the discriminability measure.
Specifically, for each image in VOC2007 val, the saliency map of each class is computed and the point with max value is extracted.
If the point falls into the annotated bounding box of the corresponding class, a hit is counted, otherwise a miss is counted.
The final accuracy is defined by considering all images and all classes therein:
\begin{equation}
    Acc (\text{all classes in all images}) = \frac{Hits}{Hits+Misses}.
    \label{pointinggame}
\end{equation}
Table \ref{acc} shows the results of representative CAM methods on discriminability evaluation.
It can be seen the gradient-based methods %\cite{gradcam,gradcamplusplus,Xgradcam,jiang2021layercam} 
are naturally more discriminative than score-based methods
%\cite{scorecam,desai2020ablation} 
as the gradients may still be prominent even for the low probability classes.
Our FD-CAM outperforms both kinds of methods by combining them in a proper manner.

\begin{table}[t]
\centering
\footnotesize
\vspace{-2pt}
\caption{Ablation study on grouped channel switching.
$\text{FD-CAM}_{ng+off}$ represents without channel grouping and only switch-off score is considered.
$\text{FD-CAM}_{g+off}$ represents with channel grouping and only switch-off is considered.
}
\label{ablation_group}
\vspace{-3pt}
\begin{tabular}{@{}lccc@{}}
\hline
Methods              & Insertion $\uparrow$     & Deletion $\downarrow$        & Overall $\uparrow$       \\ 
\hline
$\text{FD-CAM}_{ng+off}$           & 0.5260          & 0.1084          & 0.4176  
 \\
$\text{FD-CAM}_{g+off}$           & 0.5396          & 0.1020          & 0.4376  
 \\
% FD-CAM           & \textbf{0.5418}          & \textbf{0.1009}          & \textbf{0.4409} \\
FD-CAM            & \textbf{0.5534}          & \textbf{0.1001}          & \textbf{0.4533} \\
\hline
\end{tabular}
\vspace{-2pt}
\end{table}

\begin{table}[t]
\centering
\footnotesize
\vspace{-2pt}
\caption{Ablation study on weight combination schemes.
The normalized score-based weight $\hat{s}_{c}^{k}$ is used as a linear or exponential scaling factor for the gradient-based weight $\hat{\alpha}_{c}^{k}$.
The bias parameter $b$ is set to 0 or 0.5.
}
\label{ablation_weight}
\vspace{-3pt}
\begin{tabular}{@{}lccc@{}}
\hline
Methods              & Insertion $\uparrow$     & Deletion $\downarrow$        & Overall $\uparrow$       \\ 
\hline
$\text{FD-CAM}(\rho = \hat{\alpha}_{c}^k \hat{s}_{c}^{k})$           & 0.5465         & 0.1048          & 0.4417
 \\
$\text{FD-CAM}(\rho = \hat{\alpha}_{c}^k e^{\hat{s}_{c}^{k}})$           & 0.5437         & 0.1044      & 0.4393
 \\
 $\text{FD-CAM}(\rho = \hat{\alpha}_{c}^k e^{\hat{s}_{c}^{k}} - 0.5)$           & \textbf{0.5534}          & \textbf{0.1001}          & \textbf{0.4533} \\
\hline
\end{tabular}
\vspace{-6pt}
\end{table}

\subsection{Ablation Studies}
\label{sec:ablation}
\textbf{Grouped channel switching.}
In FD-CAM, we propose the grouped channel switching to perturb the channels with similar features and introduce the switch-on score to obtain the enhanced score $s_c^{k}$.
To verify the effectiveness of these two operations, we perform ablation studies and evaluate the AUC values for different versions of FD-CAM.
From Table \ref{ablation_group}, it can be seen the channel grouping and the switch-on score can both improve the performance and the full version FD-CAM can produce the best AUC values which means the explanation is more faithful.
Note, when no channel grouping and only switch-off score is applied, the FD-CAM reduces to the similar formulation of Ablation-CAM \cite{desai2020ablation}.

\textbf{Weight combination schemes.}
Similarly, we also compare different schemes for combining the gradient and score based weights and evaluate the effectiveness of the bias parameter $b$.
From Table \ref{ablation_weight}, the proposed weighting scheme can achieve the best performance.
It should be noted that we only propose one possible solution for combining the gradient and score based weights to improve CAM-based visual explanation.
We leave the investigation for better weight combination schemes to the future work.

%% file: latex/5_conclusion.tex
\section{Conclusion}

% In this paper, we propose a new CAM-based interpretable method called FD-CAM. As a new interpretable method for Convolutional Neural Networks, which combines gradients with weights generated by perturbating the network to combine convolutional activations to produce heat maps. We also discuss several properties that a good interpretable method should satisfy, and experimentally compare FD-CAM with some SOTA CAM-based techniques, the results demonstrate that FD-CAM does excellent in localization, differentiation and fidelity. In addition, we used it to mine the features of model concerns and clarify its significance in the development of deep learning. In the future, we will conduct more research and discussion on the metrics of explanations.

In this paper, we propose the FD-CAM, a novel activation map weighting scheme that shares the advantages of both gradient and score based CAM methods in producing faithful and discriminative visual explanation of CNNs.
Extensive qualitative and quantitative evaluations have shown superiority of FD-CAM and ablation studies also verify the effectiveness of the proposed modules.
% Though FD-CAM shares the advantages of both gradient and score based methods, the current weight combination scheme is still heuristic-driven.
We believe our FD-CAM can inspire more follow-ups such as learning a weight combination scheme instead of current heuristic-based formulation and designing more sophisticated channel grouping and scoring functions for more effective feature perturbation. 

 \clearpage